\documentclass[runningheads]{llncs}
\usepackage[T1]{fontenc}
\usepackage{graphicx}
\usepackage{color}
\usepackage{colortbl}
\usepackage{wrapfig}
\usepackage{caption}
\definecolor{DBlue}{RGB}{0, 51, 102} 
\definecolor{DRed}{RGB}{153, 0, 0}   
\begin{document}
\title{Improving Clinical Note Generation from Complex Doctor-Patient Conversation}
%
%
\author{Yizhan Li\inst{1} \and
Sifan Wu\inst{1} \and Christopher Smith\inst{2} \and Thomas Lo\inst{2} \and Bang Liu\inst{1}}
\authorrunning{Y. Li et al.}
\institute{Université de Montréal \& Mila - Quebec AI \\ \email{\{yizhan.li,sifan.wu,bang.liu\}@umontreal.ca}
\and GoodLabs Studio \\
\email{\{csmith,tlo\}@goodlabs.studio}}

%

%
\maketitle              
\begin{abstract}
Writing clinical notes is a critical task for healthcare professionals, serving as a vital component of patient care documentation. However, manually writing these notes is time-consuming and can impact the amount of time clinicians can spend on direct patient interaction. In this paper, we present three key contributions to the field of clinical note generation using large language models (LLMs). First, we introduce CliniKnote, a comprehensive dataset consisting of 1,200 complex doctor-patient conversations paired with their clinical notes. This dataset, created and curated by medical experts with the help of modern neural networks, provides a valuable resource for training and evaluating models in clinical note generation tasks. Second, we propose the K-SOAP (Keyword, Subjective, Objective, Assessment, and Plan) note format, which enriches traditional SOAP~\cite{podder2023soap} (Subjective, Objective, Assessment, and Plan) notes by adding a keyword section at the top, allowing for quick identification of essential information. Meanwhile, these keyword information will be fed into LLMs as guidance for generating more accurate and comprehensive notes. Third, we develop an automatic pipeline to generate K-SOAP notes from doctor-patient conversations and benchmark modern LLMs using various metrics. Our results demonstrate significant improvements in efficiency and performance compared to standard LLM finetuning methods. Demo page: \url{https://github.com/catalwaysright/K-SOAP-GOODLABS.git}
\end{abstract}

\section{Introduction}
The generation of clinical notes is a critical task for healthcare professionals, serving as a vital component of patient care documentation. However, manually dealing with these notes is time-consuming, often taking clinicians 20 to 103 min per day authoring notes and 7 to 56 min per day viewing notes\cite{10.1136/jamia.2010.008441}. This extensive time requirement may impact the amount of time clinicians can spend on direct patient interaction and other tasks. Thus, our goal is to reduce the time for both writing and viewing the notes.  Recent advancements in large language models (LLMs) have significantly enhanced text summarization capabilities, offering the potential to streamline the creation of clinical notes. These models can assist clinicians in efficiently drafting their notes, allowing them to focus more on patient care and manage multiple tasks during consultations~\cite{knoll2022userdriven}. \\
\begin{wrapfigure}{}{9cm}

    \centering
    \includegraphics[width=0.7\linewidth]{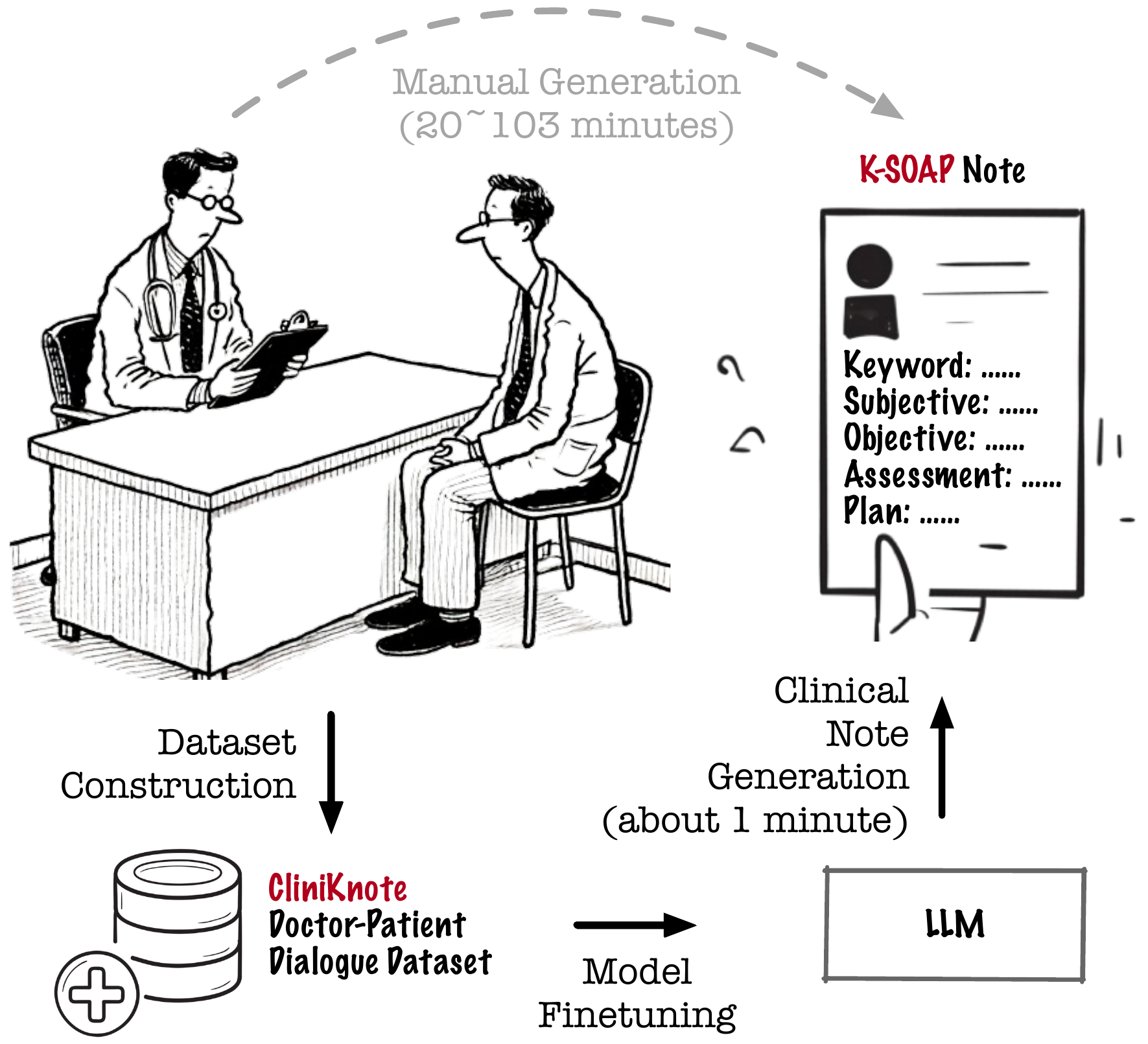}
    \caption{Demonstration of the K-SOAP note generation pipeline}
    \label{fig:intro}
\end{wrapfigure}
Despite this promise, the application in clinical note generation faces several challenges. Firstly, the quality of existing datasets is a major concern. While there are publicly available datasets like MTS-DIALOG ~\cite{ben-abacha-etal-2023-empirical} and ACI-BENCH~\cite{yim2023acibench}, they often fall short in terms of the complexity and length required to accurately simulate real-life clinical interactions. Meanwhile, we have LLM-generated dialogue datasets like Note-Chat~\cite{Wang_2024}, which do not have real medical experts involved in generating process. Secondly, the clinical note of complex doctor-patient conversations can be time-consuming for clinicians to review. This increased complexity results in longer notes with dispersed critical information, complicating the decision-making process~\cite{davidoff2011delivering}. Lastly, there is a notable need of comprehensive benchmarking for LLM-based clinical note generation, hindering the evaluation and comparison of different models and approaches in this domain. Addressing these challenges requires finetuning LLMs with domain-specific knowledge, developing high-quality datasets that reflect real-life clinical dialogues, and establishing thorough benchmarking protocols to advance the efficacy and reliability of automated clinical note generation systems~\cite{zhou2024survey}.\\
Our work makes several contributions to the field of clinical note generation using large language models. First, we introduce CliniKnote, a comprehensive dataset of complex doctor-patient conversations paired with full clinical notes and keywords. Second, we propose the K-SOAP note format, enriching traditional SOAP notes with a keyword section to improve information retrieval. Third, we perform thorough benchmarking of various large language models on two different but connected tasks (note generation and keyword detection) and develop a novel method that integrated notes with keywords and significantly improves over standard LLM finetuning approaches.\\
All conversations in CliniKnote are simulated and recorded by medical experts to ensure they closely resemble real-life doctor-patient interactions, and the paired notes are also written by real medical experts. We build the \textbf{K-SOAP} note format, which adds a keyword section at the top of traditional SOAP notes. This addition allows doctors to quickly recall domain knowledge at a glance and provides great supplement when enhancing LLM note generation as an element-aware summarization~\cite{wang2023elementawaresummarizationlargelanguage}. To achieve this, we combine clinical Named Entity Recognition (NER) and Relation Extraction (RE) to extract entities such as symptoms and diseases from the dialogue along with their relationships to the patient (e.g., present or absent). Previous clinical NER datasets like NCBI-disease~\cite{Dogan2014NCBIDC} and BC5CDR~\cite{Li2016BioCreativeVC} only contain raw disease entities without relational information. There are like some similar works like clinical negation detection~\cite{10.1093/jamia/ocaa001}, which contains affirmative, negative, speculated and recommended clinical named entities, while in CliniKnote, more types of relations with named entities are introduced.\\
Furthermore, we finetune various modern large language models and develop an automatic pipeline to generate K-SOAP notes from the conversations. The workflow of this pipeline is shown in figure \ref{fig:intro}. We benchmark CliniKnote using various automatic evaluation to assess the quality of the generated notes and their practical applicability in real-life clinical settings. We also investigate the effects of data augmentation, finetuned domain knowledge, and different adapters for various section segmentations. Our results show that the proposed method significantly reduces time complexity and improves the efficiency of generating K-SOAP notes, requiring much less time compared to manual note-writing.

\section{Related Work}
Various methods have been developed for clinical note generation. For instance, Singh \cite{singh-etal-2023-large} builds large-scale sequence-to-sequence models for generating notes from patient-doctor conversations. Using LLM for clinical note generation is increasingly popular~\cite{zhou2024survey}. Biswas \cite{Biswas_2024} demonstrates advanced prompting techniques to generate clinical notes using LLMs, while Grambow \cite{grambow-etal-2022-domain} introduces in-domain pre-training to enhance transformer models' clinical summarization performance. Additionally, the MEDIQA-2023 Dialogue2Note shared tasks~\cite{tang2023gersteinlab} include finetuning pre-trained dialogue summarization models and using few-shot in-context learning with GPT-4 to address medical dialogue summarization challenges.\\
There are some notable public clinical dialogue-note datasets. \textit{Dr.Summarize}~\cite{joshi2020dr}, generated from a telemedicine platform, and the dataset presented by \cite{pmlr-v149-chintagunta21a}, created using GPT-3, both follow a snippet-summary format. For well-structured SOAP medical notes, \textit{ACI-BENCH}~\cite{yim2023acibench} and \textit{PriMock57}~\cite{korfiatis2022primock57} feature role-played dialogues with complete SOAP notes but contain only 207 and 57 data points, respectively, which is insufficient for training modern large language models. \textit{MTS-DIALOG}~\cite{ben-abacha-etal-2023-empirical}, previously the largest dataset with clinical notes, includes 1,700 doctor-patient conversations (16k turns and 18k sentences) and summarized clinical notes (6k sentences). However, \textit{MTS-DIALOG} consists of segmented snippets and paired notes that include only some sections rather than full notes.\\
Named Entity Recognition (NER) and Relation Extraction (RE) are essential for extracting structured information from unstructured text in natural language processing. Many datasets in both fields are publicly available, recorded by BigBIO~\cite{fries2022bigbioframeworkdatacentricbiomedical}.  Recent advancements in deep learning, particularly with BiLSTM-CRF and transformer-based models like BERT, have significantly improved accuracy in identifying entities across various domains, including the medical field. Notably, recent work combining NER and RE for extracting complex information from scientific texts~\cite{dunn2022structured} and utilizing label-supervised Llama finetuning~\cite{li2023label} has achieved state-of-the-art performance. Building on these advancements, our work integrates NER and RE with large language models to enhance clinical note generation from doctor-patient dialogues.\\

\section{CliniKnote Dataset Assessment}
In this section, we discusse how we synthesize the conversations, SOAP notes, and keyword sections to create our CliniKnote dataset. We also highlight CliniKnote's advantages in terms of scale and structure by comparing with the previous largest public available note dataset MTS-DIALOGUE~\cite{ben-abacha-etal-2023-empirical}.
\begin{wrapfigure}{}{0.4\textwidth}
    \includegraphics[width=0.35\textwidth]{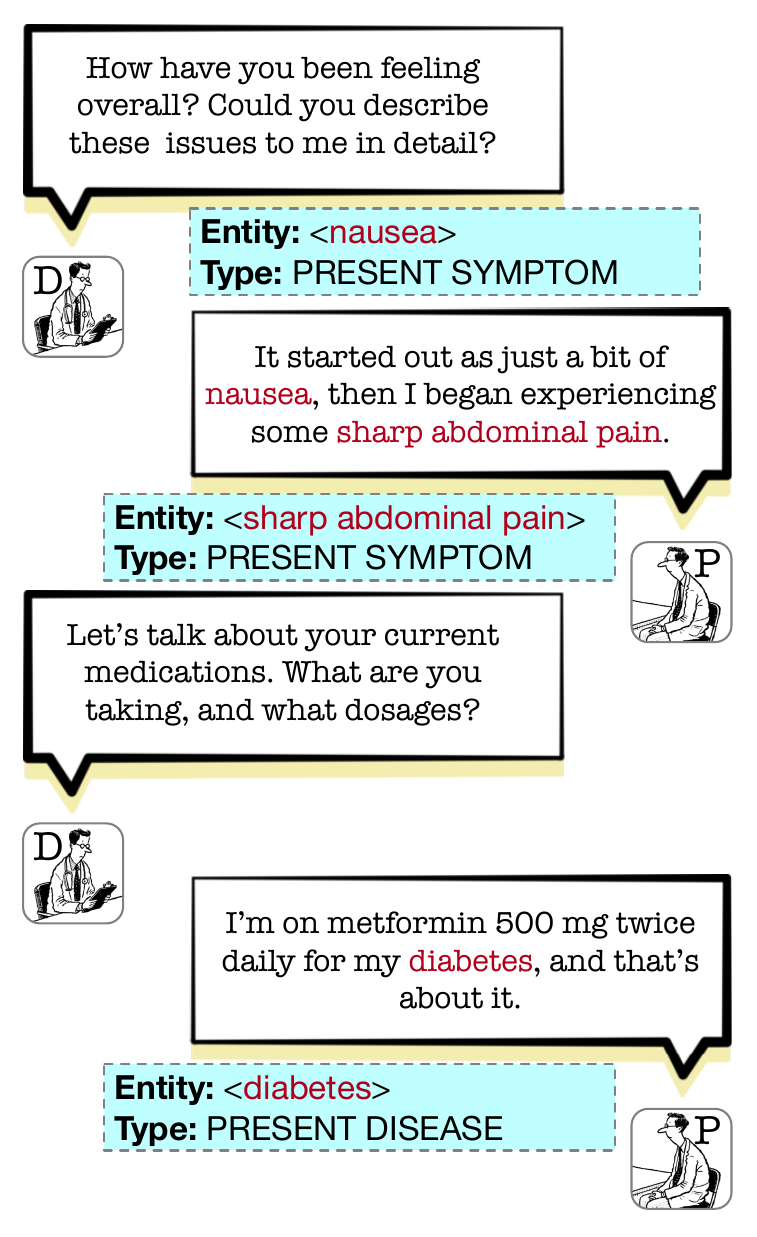}
    \caption{Keyword extraction in clinical dialogue.}
    \label{fig:keyword_demo}
\end{wrapfigure}
\subsection{Data Synthesis}
CliniKnote's dialogue and SOAP note sections are curated in a similar process as described in ~\cite{Fareez2022ADO}. A team of composed of senior medical students, medical residents, and medical doctors conducted simulated medical conversations. Following the completion of each simulated medical exam, the individual assuming the role of the physician would write a standard SOAP clinical note summarizing the findings, assessment, and medical plan of the simulated visit. Because each exam was simulated, there is no personal health information included in the dataset. The audio of the conversation was initially transcribed using automatic speech recognition (ASR) and then manually validated and corrected to minimize word error rates (WER). Through manual validation the WER of the transcripts were estimated to be approximately 1\%.\\
CliniKnote's keyword section contains all the symptoms and diseases mentioned in the conversation together with their relations to the patients labeled by GPT-4. Each keyword is prefixed to indicate its relation to the patient. Through the keyword section, we can directly find the useful symptom and disease information without looking at the full medical note. We have 8 types of entities in keyword sections in total, which are PRESENT SYMPTOM, PRESENT DISEASE, ABSENT SYMPTOM, ABSENT DISEASE, FAMILY DISEASE, PAST DISEASE, UNKNOWN DISEASE and UNKNOWN SYMPTOM respectively. The prefix PRESENT and ABSENT describe the relation between this entity and the patient. The prefix FAMILY and PAST mean that the patient has a family history of this entity while the prefix UNKNOWN means we are not sure this entity is present or absent currently and some further tests may need to be done. Figure \ref{fig:keyword_demo} shows a demo of keyword section.

\subsection{Data Augmenting}
During the development of medical exams, the simulations covered a limited number of allergies, and the occurrence of penicillin is 82.6\% among the whole dataset. Thus the dataset lacks diversity of the allergy field. Using vanilla CliniKnote for note generation might cause identical allergy generation ignoring the specific patient input.
\begin{table*}[ht]
    \centering
    \resizebox{0.9\textwidth}{!}{%
    \begin{tabular}{l|rrr|rr|rrr|rr}
        \hline
        & \multicolumn{5}{c|}{CliniKnote} & \multicolumn{5}{c}{MTS-DIALOG} \\ \hline
        & \multicolumn{3}{c|}{Dialogue} & \multicolumn{2}{c|}{Summary} & \multicolumn{3}{c|}{Dialogue} & \multicolumn{2}{c}{Summary} \\ \cline{2-11}
        & Turns & Sentences & Words & Sentences & Words & Turns & Sentences & Words & Sentences & Words \\ \hline
        count & 70,557 & 117,131 & 1,144,432 & 54,571 & 586,052 & 15,969 & 18,406 & 241,685 & 5,870 & 81,299 \\
        mean & 59 & 98 & 954 & 45 & 490 & 9 & 11 & 142 & 3 & 48 \\
        max & 102 & 155 & 1,403 & 70 & 796 & 103 & 136 & 1,951 & 57 & 1,182 \\
        min & 40 & 50 & 700 & 10 & 371 & - &-&-&-&- \\ \hline
    \end{tabular}%
    }
    \caption{Data statistics of CliniKnote and MTS-DIALOG.}
    \label{tab:comparison}
    
\end{table*}
In order to enhance the data diversity of CliniKnote, we augment the allergy section of CliniKnote. To be more specific, we first generate a list of 30 common allergies suggested by Food Standard Agency(FSA) and pick a number from 0 to 4 randomly. Then we sample that number of allergies from the list, feed those allergies and the original dialogue into GPT-4, and prompt GPT-4 to replace the original allergies with the allergies we feed and maintain the consistency of context. After replacing the allergies, we do a string comparison to make sure that is the only modified part in the dialogue.  Through data augmentation for the allergy part, we have a more complementary version of CliniKnote, which could be generalized for more clinical-related tasks.

\subsection{Data Statistics}
The final CliniKnote we introduced has 1000 dialogues in training set and 200 dialogues in test set, both paired with the proposed K-SOAP medical notes including the sections of Chief Complaint, History Of Presenting Illness, Past Medical History, Past Surgical History, Family History, Allergies, Social History, Medication, Immunization History, Review of Systems, Vital Signs, Physical Exam, Diagnostic, Assessment, Plan, and keywords.\\
\begin{wraptable}{r}{4cm}
    \centering
    \resizebox{0.3\textwidth}{!}{%
    \begin{tabular}{l|c}
        \hline
        \textbf{Entity Type} & \textbf{Count} \\ \hline
        PRESENT SYMPTOM & 12034 \\ \hline
        PRESENT DISEASE & 2442 \\ \hline
        ABSENT SYMPTOM & 24687 \\ \hline
        ABSENT DISEASE & 283 \\ \hline
        FAMILY DISEASE & 1922 \\ \hline
        PAST DISEASE & 1001 \\ \hline
        UNKNOWN DISEASE & 1538 \\ \hline
        UNKNOWN SYMPTOM & 1270 \\ \hline
    \end{tabular}
    }
    \caption{Entity counts across different categories.}
    \label{tab:entity_counts}
\end{wraptable}
Compared with MTS-DIALOG, CliniKnote has more data points and more complex dialogues that are closer to real-life clinical conversations. Table \ref{tab:comparison} shows the comparison between CliniKnote and MTS-DIALOG, demonstrating our large scales in word count and turns. Word count is 5 times than that of MTS-DIALOG in dialogues and 7 times in summary. The final keyword section contains 12,286 short dialogues segmented from the original dialogues paired with the entities and relation labeled. Table \ref{tab:entity_counts} shows the specific count for each type of entity.

\section{K-SOAP Generation Based On Medical Dialogue}
The complete task aims to generate K-SOAP note based on the input dialogue. We divide this task into two parts. The first part is to to do keyword extraction in NER format and the second is to generate the traditional SOAP note with the guide of keywords in two complex sections. The concatenation of these two parts made up to the final K-SOAP note.

\subsection{Keyword Extraction}
We generate the keyword section for each clinical dialogue to reduce the reviewing complexity and provide guidance for note generation. We treat this task as a NER task but with relation as the prefix. For keyword extraction, we use Label-supervised unmasked Llama finetuning~\cite{li2023label}. Instead of autoregressive generation, we add a label predictor at the top of a Llama-7b model. The dialogue sequences decoded by Llama will be fed into this predictor to predict the labels. We also keep the idea of removing the causal masks because their empirical studies show that using token representations learned with causal masks significantly underperforms in token classification tasks. We merge the diseases and symptoms together into only 4 types, which are PRESENT, ABSENT, FAMILY and UNKNOWN. 
To be more specific, we set the word to tag dictionary as {{`O': 0, `B-PRESENT': 1, `B-ABSENT': 2, `B-FAMILY': 3, `B-UNKNOWN': 4, `I-PRESENT': 5, `I-ABSENT':6, `I-FAMILY':7, `I-UNKNOWN':8}} while B means start and I means inside. The model will determine whether a word is a valid entity and what is the relation with the patient at the same time.

\subsection{SOAP Clinical Note Generation}
The main task for our CliniKnote dataset is to generate the traditional SOAP clinical notes. To compare the performance of models with different sizes, domain knowledge, and released versions, we selected the following public models to finetune for our benchmarking: qCammel-13b \cite{qCammel}, a finetuned version of the Llama-2 13B model, trained on a distilled dataset of 15,000 medical domain instructions using Quantized Low-Rank 
Adaptation(QLoRA)~\cite{dettmers2023qlora}; 
Llama2-13b \cite{touvron2023Llama}; 
Llama3-8b \cite{Llama3modelcard}; 
and OpenBioLLM-Llama3-8B \cite{OpenBioLLMs}, 
which is finetuned on the biomedical domain. Additionally, we include commercial models such as GPT-3.5-turbo and GPT-4 for our benchmarking. We use offical QLoRA to finetune all the large language models. For the supervised learning, there are two versions of the input. The first consists of a prompt filled with the original dialogue, while the second enhances this prompt by incorporating keyword information. The output is the clinical note in CliniKnote dataset reformatted in json mode as it will be more convenient to access. After finetuning, the Parameter-Efficient finetuning (PEFT) adapters~\cite{peft} will be saved and loaded to do further evaluation.

\section{Evaluation of K-SOAP Note Generation}

We evaluate CliniKnote using various metrics to assess note quality and time efficiency, highlighting its effectiveness in producing high-quality clinical notes and enhancing medical workflows.

\subsection{Evaluation Metrics}
The evaluation of clinical note generation is particularly challenging, as it is difficult to accurately measure the relevance and quality of the generated notes due to the complexity and specificity of medical information.\\
For keyword evaluation, we only record and compare the accuracy, precision, recall and F1 score. For SOAP part note evaluation, we employ several metrics to ensure a comprehensive assessment. We start with the most traditional metric, ROUGE~\cite{Lin2004ROUGEAP}, which evaluates similarity based on common subsequences between the generated and reference texts. In addition to ROUGE, we introduce more advanced metrics: BERTScore~\cite{bert-score} and BLEURT ~\cite{sellam2020bleurt} ~\cite{pu2021learning}. These metrics leverage contextual embeddings from BERT to evaluate the semantic similarity between generated and reference texts. BERTScore uses token embeddings to compute precision, recall, and F1 scores, providing a more nuanced and context-aware evaluation compared to traditional metrics. Similarly, BLEURT incorporates pre-trained transformer models to better capture the meaning and fluency of the text, offering insights into the semantic quality of the generated notes. We use Roberta-large model for BERTScore metric and the newest checkpoint BLEURT-20 for BLEURT metric. We also include QuestEval~\cite{scialom2020QuestEval} in our evaluation framework. Unlike traditional metrics, QuestEval does not require any ground truth references. Compared with ROUGE, QuestEval substantially improves the correlation with human judgments over four evaluation dimensions: consistency, coherence, fluency, and relevance. By incorporating question-answering models, QuestEval evaluates the generated text's ability to provide accurate and relevant information, making it particularly suitable for complex domains like clinical note generation.


\section{Experiments}
In this section, we focus on the performance of various models after finetuning on CliniKnote. We also discuss different finetuning methods for keyword extraction.
\begin{table}[h]
    \centering

    \resizebox{0.6\columnwidth}{!}{%
    \begin{tabular}{c|c|c|c|c|c}
        \hline
        \textbf{Model} & \textbf{Acc} & \textbf{F1} & \textbf{Pre} & \textbf{Recall} & \textbf{Eval\_Loss} \\ \hline
        \textbf{biobert} & \underline{0.933} & \underline{0.735} & \textbf{0.755} & \underline{0.715} & \underline{0.197} \\ \hline
        \textbf{scibert} & 0.921 & 0.669 & 0.723 & 0.622 & 0.221 \\ \hline
        \textbf{pubmedbert} & 0.919 & 0.685 & 0.689 & 0.681 & 0.230 \\ \hline
        \textbf{bluebert} & 0.903 & 0.554 & 0.679 & 0.467 & 0.285 \\ \hline
        \textbf{bert} & 0.915 & 0.653 & 0.677 & 0.630 & 0.242 \\ \hline
        \textbf{Llama2-7b-in-context} & - & 0.692 & 0.391 & 0.500 & - \\ \hline
        \textbf{Llama2-7b-QA} & - & 0.642 & \underline{0.731} & 0.657 & 0.307 \\ \hline
        \textbf{Llama2-7b-}\\ \textbf{label-supervised} & \textbf{0.971} & \textbf{0.752} & 0.726 & \textbf{0.780} & \textbf{0.107} \\ \hline
    \end{tabular}%
    }
    \caption{Keyword performance comparison of different models.}
    \label{tab:performance_comparison}
\end{table}
\subsection{Experimental Settings}
We train all the models with one A100-80GB GPU. For clinical note generation task, we train all the large language models for two epochs. We use the default setting for question-answer finetuning with QLoRA. We set the learning rate to 0.0002 and the warmup ratio to 0.03. For QLoRA hyperparameters, we set LoRA $r$ to 64 and the loRA $\alpha$ to 16 and all the models are quantized in 4 bits. For keyword task, we finetune the model by LoRA. We set batch size as 8, learning rate as 1e-4, Max length as 256, LoRA $r$ as 256, LoRA $\alpha$ as 768, LoRA dropout as 0.1. We train Llama2-7b model for 5 epochs.
\begin{table*}[hbt!]
    \centering
    \resizebox{0.9\columnwidth}{!}{%
    \begin{tabular}{|p{2cm}|p{10cm}|}
        \hline
        \textbf{Dialogue} &  
        \textbf{D:} When did you first notice this rash? \textbf{P:} It started about two weeks ago, just a small patch. \textbf{D:} Has anything in particular seemed to make it better or worse? \textbf{P:} It gets itchier at night, and when I sweat.  \\  \hline

        \textbf{Keywords} &
        \textcolor{DRed}{[`1rash', `|', `1itchier', `5at', `5night', `|', `5sweat']} \newline 
        \textcolor{DBlue}{[`1rash', `|', `1itchier', `5at', `5night']} \\ \hline

        \textbf{Dialogue} &
        \textbf{D:} Any cough or shortness of breath? \textbf{P:} No. \textbf{D:} Any changes in urination or blood in the urine? \textbf{P:} No. \textbf{D:} Any changes in bowel movements, nausea, vomiting, or abdominal pain? \textbf{P:} No. \\ \hline

        \textbf{Keywords} & \textcolor{DRed}{[`2cough', `|', `2shortness', `6of', `6breath', `|', `2changes', `6in', `6urination', `|', `2blood', `6in', `6the', `6urine', `|', `2changes', `6in', `6bowel', `6movements', `|', `2nausea', `|', `2vomiting', `|', `2abdominal', `6pain']}
         \newline 
        \textcolor{DBlue}{[`2cough', `|', `2shortness', `6of', `6breath', `|', `2changes', `6in', `6urination', `|', `2blood', `6in', `6the', `6urine', `|', `2changes', `6in', `6bowel', `6movements', `|', `2nausea', `|', `2vomiting', `|', `2abdominal', `6pain']} \\ \hline
    \end{tabular}
    }
    \caption{Comparison of Keywords Extraction in Dialogues: Red for Prediction, Blue for Ground Truth.}
    \label{tab:demo_keywords}
\end{table*}

\subsection{Keyword Extraction Results}
We use BioBart, SciBert, PubMed, BlueBert and Bert as baseline models for key words extraction task. For baseline models, we finetune the pretrained models on CliniKnote dataset. The learning rate is set to 1e-4, batch size is set to 128 and we train all baseline models for 10 epochs. Besides the label-supervised Llama finetuning, we also tried in-context learning and question-answer based finetuning on the same Llama2-7b model. Table \ref{tab:performance_comparison} shows the results among all the models and Llama2-7b with our label-supervised finetuning dominates all the metrics except for Precision and has the lowest $Eval Loss$. The precision of Llama2-7b-label-supervised is slightly lower than that of biobert. Neither in-context learning nor QA-based finetuning on Llama2-7b has decent performance on this task since it combines RE as prefix and makes it harder to predict for autoregressive LLM. Table \ref{tab:demo_keywords} shows two demos of the Llama2-7b-label-supervised's prediction and the ground truth. Label-supervised finetuning enhances the entity recognition ability of large language models while maintaining their comprehension skills.  The results also demonstrate that raw llms cannot handle this keyword extraction task with decent performance, which indicates the necessity of the training resources provided in CliniKnote keyword section.

\subsection{Clinical Note Generation Results}
In this section, all models are finetuned and tested in 3 modes, which are full note mode, section-4 mode and section-15 mode that contain different numbers of LoRA adapter~\cite{hu2021lora}. Full note mode means we only train one adapter to generation the full clinical note. Section-4 mode means we train four different adapters to generate the full note. How we divide the four parts is a little different from how we separate Subjective, Objective, Assessment, and Plan parts. We only have Chief Complaint and History Of Presenting Illness in the first part. The second part contains Past Medical History, Past Surgical History, Family History, Allergies, Social History, Medication List, Immunization History and Review of Systems. The third part contains Vital Signs and Physical Exam while the final part contains Assessment, Diagnosis and Plans. Section-15 mode means we train 15 different adapters to generate all the fifteen sections contained in the note separately. For each section, we evaluate the performance difference of our best model with and without the inclusion of keyword guidance. As for commercial models, we try full note zeroshot generation and oneshot(one demo in prompt) in-context learning. The total generation time is all below 2 min. \\
\begin{table*}[]
\small
\scalebox{0.65}{
\begin{tabular}{cccccccccc}
\hline
\textbf{Model}                                                   & \textbf{Rouge-1} & \textbf{Rouge-2} & \textbf{Rouge-L} & \begin{tabular}[c]{@{}c@{}}\textbf{Rouge-}\\ \textbf{lsum}\end{tabular} & \textbf{Bertscore-P} & \textbf{Bertscore-R} & \textbf{Bertscore-F1} & \textbf{Bleurt} & \textbf{QuestEval} \\ \hline
 \textbf{baseline models} &  & & & & & & & & \\ 
\begin{tabular}[c]{@{}c@{}}Bart-full\end{tabular}    &  0.44       &    0.29     &  0.33       &  0.417                                                     &  \textbf{0.908}                                                      & 0.808                                                       & 0.853                                                        & \underline{0.563}       & \textbf{0.498}          \\ 
\begin{tabular}[c]{@{}c@{}}Bart-section-4\end{tabular}  
  &   \textbf{0.633}       &   0.388      &  \textbf{0.459}       & \underline{0.606}                                                      & 0.869                                                       & \underline{0.887}                                                       & \textbf{0.878}                                                        & 0.518       &  0.463         \\ 
\begin{tabular}[c]{@{}c@{}}Bart-section-15\end{tabular}    &   0.591      &  \underline{0.367}       &  \underline{0.406}       &  0.572                                                     & 0.801                                                       &   0.842                                                     & 0.821                                                        & 0.372       & 0.484          \\

\begin{tabular}[c]{@{}c@{}}biobart-full\end{tabular} &  0.439       &    0.303     &  0.336       &  0.417                                                     &  \underline{0.906}                                                      &  0.810                                                      &0.855                                                         & \textbf{0.567}       & \textbf{0.506}          \\ 
\begin{tabular}[c]{@{}c@{}}biobart-section-4\end{tabular} & \textbf{0.633}        &  \textbf{0.402}       & 0.382        & \textbf{0.608}                                                      & 0.867                                                       & \textbf{0.890}                                                        & \textbf{0.878}                                                        & 0.512       & 0.459           \\ 
\begin{tabular}[c]{@{}c@{}}biobart-section-15\end{tabular} &  0.583       &  0.366       &  0.399       &   0.566                                                    &  0.784                                                      &  0.837                                                      &  0.809                                                       & 0.318       &  0.478         \\ \hline

  \textbf{open-source LLMs}  & & & & & & & & & \\
\begin{tabular}[c]{@{}c@{}}Llama2-13b-full\end{tabular}                                                    &   0.641      &  0.409       &    0.505     &   0.591                                                    &  0.871                                                      &  0.852                                                      &  0.861                                                       & 0.534       & 0.461          \\ 
\begin{tabular}[c]{@{}c@{}}Llama2-13b-section-4\end{tabular}                                                    &  \underline{0.678}       & \underline{0.437}        &   \underline{0.516}      &    \underline{0.632}                                                   &   0.871                                                     & 0.852                                                       &  0.861                                                       &  0.553      & 0.399          \\ 
\begin{tabular}[c]{@{}c@{}}Llama2-13b-section-15\end{tabular}                                                    &    0.656     &   0.426      &  0.506       &   0.613                                                    &   0.872                                                     &         0.861                                               &  0.867                                                       & 0.495       & 0.426          \\ 
\begin{tabular}[c]{@{}c@{}}qCammel-13b-full\end{tabular}                                                    &  0.639       &  0.395       &   0.477      &    0.568                                                   &   0.879                                                     &  0.865                                                      &   0.872                                                      &  \underline{0.581}      &  \underline{0.475}         \\ 
\begin{tabular}[c]{@{}c@{}}qCammel-13b-section-4\end{tabular}                                                    & 0.465        &  0.289       & 0.354        &     0.426                                                  &   0.889                                                     & 0.866                                                       &  0.877                                                       &  0.476      &      0.327     \\ 
\begin{tabular}[c]{@{}c@{}}qCammel-13b-section-15\end{tabular}                                                    &  \textbf{0.692}       &   \textbf{0.467}      &  \textbf{0.538}       &   \textbf{0.646}                                                    &  \textbf{0.896}                                                      &      \textbf{0.893}                                                  &   \textbf{0.895}                                                      &  0.474      &   0.395        \\ 
\begin{tabular}[c]{@{}c@{}}Llama3-8b-full\end{tabular}                                                    &  0.641       &  0.409       &    0.505     &   0.591                                                    &  \underline{0.893}                                                      &  \underline{0.878}                                                      &  \underline{0.885}                                                       & \textbf{0.595}       &  \textbf{0.476}         \\ 
\begin{tabular}[c]{@{}c@{}}Llama3-8b-section-4\end{tabular}                                                    &  \underline{0.678}       &  \underline{0.437}       &  \underline{0.516}       &     \underline{0.632}                                                  &  0.867                                                      &  0.860                                                      &  0.864                                                       & 0.548       &  0.468         \\ 
\begin{tabular}[c]{@{}c@{}}Llama3-8b-section-15\end{tabular}                                                    &  0.656       &  0.426       &  0.506       &   0.613                                                    &   0.866                                                     &  0.857                                                      &  0.862                                                       & 0.550       & 0.465          \\ 
\begin{tabular}[c]{@{}c@{}}OpenBioLLM-Llama3\\-8B-full\end{tabular}                                                    &  0.372       &  0.189       & 0.254        &   0.307                                                    &  0.839                                                      &  0.798                                                      &  0.818                                                       & 0.421       & 0.421          \\ 
\begin{tabular}[c]{@{}c@{}}OpenBioLLM-Llama3\\-8B-section-4\end{tabular}                                                    &     0.386    &  0.194       &  0.253       &   0.340                                                    &   0.834                                                     &   0.816                                                     &     0.825                                                    & 0.319       &  0.364         \\ 
\begin{tabular}[c]{@{}c@{}}OpenBioLLM-Llama3-8B\\-section-15\end{tabular}                                                    &     0.298    &  0.207       &   0.248      &  0.282                                                     &   0.834                                                     &    0.816                                                    &  0.825                                                       & 0.297       & 0.316          \\ \hline
 \textbf{Commercial models} & & & & & & & & & \\
\begin{tabular}[c]{@{}c@{}}GPT-3.5-Turbo-\\ zeroshot\end{tabular}                                                    &  0.592       &  0.322       &   0.441      &    0.562                                                   &    0.802                                                    &   0.846                                                     & 0.823                                                        & 0.500       & 0.407          \\ 
\begin{tabular}[c]{@{}c@{}}GPT-3.5-Turbo-\\ oneshot\end{tabular}                                                    & \underline{0.705}        & \underline{0.451}        &  \underline{0.580}       & \underline{0.663}                                                      &  \underline{0.855}                                                      &     \underline{0.889}                                                   &  \underline{0.871}                                                       & \underline{0.560}       & \underline{0.441}          \\ 
\begin{tabular}[c]{@{}c@{}}GPT-4o-\\ zeroshot\end{tabular}                                                     &  0.630       &  0.363       &  0.491       &  0.602                                                     & 0.816                                                       & 0.847                                                       & 0.831                                                        & 0.467       &   0.405        \\ 
\begin{tabular}[c]{@{}c@{}}GPT-4o-\\ oneshot\end{tabular}                                                     &   \textbf{0.753}      &  \textbf{0.519}       &   \textbf{0.634}      &   \textbf{0.704}                                                    &  \textbf{0.878}                                                      &  \textbf{0.899}                                                      & \textbf{0.889}                                                        & \textbf{0.610}       & \textbf{0.493} \\ \hline    
\end{tabular}}
\caption{Overall results for medical dialogue to note task on CliniKnote dataset.}
\label{tab:main_result}
\end{table*}
We choose Bart and Biobert as baseline models.For matching clinical tasks, we use the continued pre-training on PubMed abstract model of BART-large version. We further compare the performance of BioBart, which has the same model structure as BART but possesses distinct tokenizers and vocabulary size.\\
Table \ref{tab:main_result} shows the performance of all the models with the automatic metrics we pick. qCammel-13b-section-15 achieves overall best performance on ROUGE and BERTScores among all open-scouce models while GPT-4o-oneshot is considered the best among all commercial models. Section-15 mode does not show significant enhancement on the performance even though it takes 11 and 14 more adapters than other modes. All modes of Llama3-8b show decent results. The section-4 mode of Llama3-8b shows second best peformance on ROUGE score and full mode has highest score on Bleurt, QuestEval and second highest score on Bertscore among all open-source models. However, OpenBioLLM-Llama3-8b has poor performance on almost all metrics. All three modes of OpenBioLLM-Llama3-8b shows lowest scores on ROUGE, BertScore, Bleurt and QuestEval, even lower than baseline models. The domain knowledge insertion triggers decreasing on information extraction ability. The benchmarking shows that appropriate in-domain instruction is beneficial while it could also jeopardize the model's original extraction ability. Table \ref{tab:qCammel_demo} shows some sections generated by qCammel-13b-section-15 and compares them with the reference sections. After data augmenting, the occurrence rate of penicillin as an allergy drops by 11.5\% The actual occurrence rate of penicillin is 19.5\% in test set. However, the qCammel-13b-section-15 model predicts penicillin as an allergy 31.5\% of the time without data augmentation. Specifically, the model tends to output penicillin when no allergy is mentioned in the dialogue. After applying data augmentation, the penicillin prediction rate drops to 20\%, which aligns much closer to the real distribution.\\
\begin{table*}[h]
    \centering
    \resizebox{0.9\columnwidth}{!}{%
    \begin{tabular}{|p{0.4cm}|p{5.2cm}|p{5.2cm}|}
        \hline
        & \textbf{Generated Section} & \textbf{Reference Section} \\ \hline
        1 & {"Chief Complaint": "Skin rash on arms and legs."} 
          & {"Chief Complaint": "Skin issues on forearm."} \\ \hline
        2 & {"History Of Presenting Illness": "62-year-old female presents with a skin issue that started as a small red patch on her forearm about a month ago. The patch has been getting larger and itchier, rated as a 7 out of 10 in severity. Cold compresses provide some relief, while heat and wearing long sleeves aggravate the condition. The itching is affecting her sleep and self-esteem."}
          & {"History Of Presenting Illness": "62-year-old female presents with a skin issue that started about a month ago. Initially a small red patch on the forearm, which has been getting larger and itchier. Itchiness rates at about a 7 out of 10. Symptoms worsen at night and when wearing long sleeves. Cold compresses provide slight relief, while heat and fabric contact aggravate the condition. The issue has impacted sleep and caused self-consciousness about appearance. No prior occurrence of similar symptoms. Over-the-counter creams have been tried with little effect. "} \\ \hline
        3 & {'Family History': ['Father had psoriasis, mother has a history of eczema. No other major illnesses that I'm aware of.']}
          & {'Family History': ['Father had psoriasis, mother has eczema.']} \\ \hline
        4 & {'Allergies': ['Latex: severe skin irritation']}
          & {'Allergies': ['Latex - causes severe skin irritation.']} \\ \hline
        5 & {"Social History": "Patient is a retired librarian, widowed with two adult children. She enjoys gardening and has never smoked. Drinks wine occasionally with dinner."}
          & {"Social History": "Retired librarian, widowed with two adult children. Enjoys gardening, non-smoker, minimal alcohol consumption."} \\ \hline
    \end{tabular}
    }
    \caption{Examples of generated summaries by qCammel-13b-section-15.}
    \label{tab:qCammel_demo}
\end{table*}
The inclusion of keyword guidance significantly enhances note generation for complex sections, achieving a 21.5\% increase in F1 score for the Assessment section and a 22.5\% increase for the History of Presenting Illness section. The addition of keywords helps guide the model to focus more effectively on the correct information. Besides the accuracy improvement of generation, we also notice obvious prevention of hallucinations, as verified through a manual review of the output differences. However, a noticeable performance drop is observed in simpler sections such as Chief Complaint and Allergies. This decline likely occurs because these sections do not rely on keyword information, which may introduce noise rather than value.
\subsection{Conclusion}
This study introduces automatic clinical note generation to improve medical documentation efficiency and accuracy with the guidance of keywords. We introduce the K-SOAP format, adding keywords to traditional SOAP notes for quicker review, and CliniKnote, a large-scale dataset of complex dialogues paired with K-SOAP notes. Training large language models on CliniKnote significantly improved clinical note accuracy. Our evaluation highlights its potential to streamline documentation and save doctors valuable time.


%
%
%
\bibliographystyle{splncs04}
\bibliography{splncs04}

\begin{thebibliography}{10}
\providecommand{\url}[1]{\texttt{#1}}
\providecommand{\urlprefix}{URL }
\providecommand{\doi}[1]{https://doi.org/#1}

\bibitem{Llama3modelcard}
AI@Meta: Llama 3 model card (2024)

\bibitem{OpenBioLLMs}
Ankit~Pal, M.S.: Openbiollms: Advancing open-source large language models for healthcare and life sciences. \url{https://huggingface.co/aaditya/OpenBioLLM-Llama3-70B} (2024)

\bibitem{ben-abacha-etal-2023-empirical}
Ben~Abacha, A., Yim, W.w., Fan, Y., Lin, T.: An empirical study of clinical note generation from doctor-patient encounters. Association for Computational Linguistics (May 2023), \url{https://aclanthology.org/2023.eacl-main.168}

\bibitem{Biswas_2024}
Biswas, A., Talukdar, W.: Intelligent clinical documentation: Harnessing generative ai for patient-centric clinical note generation p. 994–1008 (May 2024), \url{http://dx.doi.org/10.38124/ijisrt/IJISRT24MAY1483}

\bibitem{pmlr-v149-chintagunta21a}
Chintagunta, G.e.a.: Medically aware gpt-3 as a data generator for medical dialogue summarization. pp. 354--372. PMLR (06--07 Aug 2021), \url{https://proceedings.mlr.press/v149/chintagunta21a.html}

\bibitem{davidoff2011delivering}
Davidoff, F., Miglus, J.: Delivering clinical evidence where it's needed: building an information system worthy of the profession. Jama  \textbf{305}(18),  1906--1907 (2011)

\bibitem{dettmers2023qlora}
Dettmers, T., Pagnoni, A., Holtzman, A., Zettlemoyer, L.: Qlora: Efficient finetuning of quantized llms (2023)

\bibitem{Dogan2014NCBIDC}
Dogan, R.I., Leaman, R., Lu, Z.: Ncbi disease corpus: A resource for disease name recognition and concept normalization. Journal of biomedical informatics  \textbf{47},  1--10 (2014), \url{https://api.semanticscholar.org/CorpusID:234064}

\bibitem{dunn2022structured}
Dunn, A., Dagdelen, J., et~al.: Structured information extraction from complex scientific text with fine-tuned large language models (2022)

\bibitem{Fareez2022ADO}
Fareez, F., Parikh, T., et~al.: A dataset of simulated patient-physician medical interviews with a focus on respiratory cases. Scientific Data  \textbf{9} (2022), \url{https://api.semanticscholar.org/CorpusID:249747721}

\bibitem{fries2022bigbioframeworkdatacentricbiomedical}
Fries, J.A., Weber, L., Seelam, N., et~al.: Bigbio: A framework for data-centric biomedical natural language processing (2022), \url{https://arxiv.org/abs/2206.15076}

\bibitem{grambow-etal-2022-domain}
Grambow, C., Zhang, L., Schaaf, T.: In-domain pre-training improves clinical note generation from doctor-patient conversations. Association for Computational Linguistics (Jul 2022), \url{https://aclanthology.org/2022.nlg4health-1.2}

\bibitem{10.1136/jamia.2010.008441}
Hripcsak, G., Vawdrey, D.K., Fred, M.R., Bostwick, S.B.: Use of electronic clinical documentation: time spent and team interactions. Journal of the American Medical Informatics Association  \textbf{18}(2),  112--117 (02 2011), \url{https://doi.org/10.1136/jamia.2010.008441}

\bibitem{hu2021lora}
Hu, E.J., Shen, Y., Wallis, P., Allen-Zhu, Z., Li, Y., Wang, S., Wang, L., Chen, W.: Lora: Low-rank adaptation of large language models (2021)

\bibitem{joshi2020dr}
Joshi, A., Katariya, N., Amatriain, X., Kannan, A.: Dr. summarize: Global summarization of medical dialogue by exploiting local structures (2020)

\bibitem{knoll2022userdriven}
Knoll, T., Moramarco, F., et~al.: User-driven research of medical note generation software (2022)

\bibitem{korfiatis2022primock57}
Korfiatis, A.P., Moramarco, F., Sarac, R., Savkov, A.: Primock57: A dataset of primary care mock consultations (2022)

\bibitem{Li2016BioCreativeVC}
Li, J., Sun, Y., et~al.: Biocreative v cdr task corpus: a resource for chemical disease relation extraction. Database: The Journal of Biological Databases and Curation  \textbf{2016} (2016), \url{https://api.semanticscholar.org/CorpusID:88817}

\bibitem{li2023label}
Li, Z., Li, X., et~al.: Label supervised llama finetuning (2023)

\bibitem{10.1093/jamia/ocaa001}
Lin, C., Bethard, S., Dligach, e.a.: Does bert need domain adaptation for clinical negation detection? Journal of the American Medical Informatics Association  \textbf{27}(4),  584--591 (02 2020), \url{https://doi.org/10.1093/jamia/ocaa001}

\bibitem{Lin2004ROUGEAP}
Lin, C.Y.: Rouge: A package for automatic evaluation of summaries. In: Annual Meeting of the Association for Computational Linguistics (2004), \url{https://api.semanticscholar.org/CorpusID:964287}

\bibitem{peft}
Mangrulkar, S., Gugger, S., et~al.: Peft: State-of-the-art parameter-efficient fine-tuning methods. \url{https://github.com/huggingface/peft} (2022)

\bibitem{podder2023soap}
Podder, V., Lew, V., Ghassemzadeh, S.: SOAP Notes. StatPearls Publishing, jan- edn. (2024), \url{https://www.ncbi.nlm.nih.gov/books/NBK482263/}, [Updated 2023 Aug 28]

\bibitem{pu2021learning}
Pu, A., Chung, H.W., Parikh, A.P., Gehrmann, S., Sellam, T.: Learning compact metrics for mt. In: Proceedings of EMNLP (2021)

\bibitem{scialom2020QuestEval}
Scialom, e.a.: Questeval: Summarization asks for fact-based evaluation. arXiv preprint arXiv:2103.12693  (2021)

\bibitem{sellam2020bleurt}
Sellam, T., Das, D., Parikh, A.P.: Bleurt: Learning robust metrics for text generation. In: Proceedings of ACL (2020)

\bibitem{singh-etal-2023-large}
Singh, G., Pan, e.a.: Large scale sequence-to-sequence models for clinical note generation from patient-doctor conversations (Jul 2023), \url{https://aclanthology.org/2023.clinicalnlp-1.18}

\bibitem{tang2023gersteinlab}
Tang, X., Tran, A., Tan, J., Gerstein, M.: Gersteinlab at mediqa-chat 2023: Clinical note summarization from doctor-patient conversations through fine-tuning and in-context learning (2023)

\bibitem{qCammel}
Toma, A.: qcammel on huggingface (2023), \url{https://huggingface.co/augtoma/qCammel-13}

\bibitem{touvron2023Llama}
Touvron, H., Martin, L., Stone, K., et~al.: Llama 2: Open foundation and fine-tuned chat models (2023)

\bibitem{Wang_2024}
Wang, J., Yao, Z., Yang, e.a.: Notechat: A dataset of synthetic patient-physician conversations conditioned on clinical notes. p. 15183–15201. Association for Computational Linguistics (2024), \url{http://dx.doi.org/10.18653/v1/2024.findings-acl.901}

\bibitem{wang2023elementawaresummarizationlargelanguage}
Wang, Y., Zhang, Z., Wang, R.: Element-aware summarization with large language models: Expert-aligned evaluation and chain-of-thought method (2023), \url{https://arxiv.org/abs/2305.13412}

\bibitem{yim2023acibench}
wai Yim, W., Fu, Y., et~al.: Aci-bench: a novel ambient clinical intelligence dataset for benchmarking automatic visit note generation (2023)

\bibitem{bert-score}
Zhang*, T., Kishore*, V., Wu*, F., Weinberger, K.Q., Artzi, Y.: Bertscore: Evaluating text generation with bert. In: International Conference on Learning Representations (2020), \url{https://openreview.net/forum?id=SkeHuCVFDr}

\bibitem{zhou2024survey}
Zhou, H., Liu, F., Gu, B., et~al.: A survey of large language models in medicine: Progress, application, and challenge (2024)

\end{thebibliography}

%




\end{document}